\documentclass{article} 
\usepackage{iclr2027_conference,times}

\usepackage{amsmath,amsfonts,bm}

\def\eqref#1{equation~\ref{#1}}

\def\1{\bm{1}}

\DeclareMathAlphabet{\mathsfit}{\encodingdefault}{\sfdefault}{m}{sl}
\SetMathAlphabet{\mathsfit}{bold}{\encodingdefault}{\sfdefault}{bx}{n}

\usepackage{hyperref}
\usepackage{url}
\usepackage{booktabs}
\usepackage{graphicx}
\usepackage{svg}

\iclrfinalcopy

\title{TIGPO: Temporal Instance-Graph Policy Optimization for
Long-Horizon LLM Agents}

\author{Jinwei Gan \\
Department of Computer Science \\
Nanjing University \\
\texttt{ganjw@smail.nju.edu.cn}
}
\begin{document}

\maketitle

\fancyhead{}
\fancyhead[L]{Under review as a conference paper at ICLR 2027}
\renewcommand{\headrulewidth}{0.4pt}




\begin{abstract}
Graph-based policy optimization improves credit assignment for long-horizon
LLM agents by organizing rollout trajectories into state-transition graphs.
However, existing methods construct graphs independently within each policy
update, discarding transitions discovered by earlier policies and limiting
advantage estimation to small, batch-local rollout groups. We propose
\emph{Temporal Instance-Graph Policy Optimization} (TIGPO), which extends
graph-based credit assignment across policy updates. TIGPO maintains a
persistent transition graph for each task, allowing valid transitions
discovered by different policy versions to jointly determine credit for
current rollouts. To actively reconnect current exploration with historical
experience, TIGPO allocates a fixed rollout budget between Exploration slots
for ordinary task sampling and Revisit slots for delayed reattempts of
previously explored tasks. For each revisit, TIGPO pairs the current rollout
group with its corresponding earlier Exploration group to construct a
cross-temporal reference. The enlarged reference is designed to stabilize
relative advantage estimation under small rollout groups, while comparison on
the same task directly captures policy improvement across training stages.
Historical transitions and scores serve only as structural and detached
statistical references and are never replayed in the policy loss. Experiments
on ALFWorld and WebShop demonstrate that TIGPO consistently outperforms prior
group-based and graph-based policy optimization methods.
\end{abstract}

\section{Introduction}

\label{sec:introduction}

Large language models (LLMs) have rapidly evolved from static text generators
into interactive agents that perceive environment states, reason over
observations, and execute actions to accomplish complex goals
\citep{yao2023react}. Representative applications include embodied agents
operating in simulated households \citep{shridhar2021alfworld} and web agents
navigating e-commerce environments \citep{yao2022webshop}. Unlike single-turn
reasoning, these tasks require an agent to make a sequence of interdependent
decisions over a long interaction horizon. Rewards are often sparse and
delayed, with success determined only after the entire trajectory is
completed. Consequently, assigning credit to individual actions remains a
central challenge for reinforcement learning (RL) of LLM agents.

Traditional actor--critic algorithms such as Proximal Policy Optimization
(PPO) estimate action advantages using a learned value function
\citep{schulman2017ppo}. More recently, Group Relative Policy Optimization
(GRPO) has provided a value-free alternative by estimating relative advantages
from groups of sampled outputs \citep{shao2024deepseekmath}. This group-based
paradigm has subsequently been extended to long-horizon agentic tasks. GiGPO
introduces episode-level and anchor-state-level groups to obtain more
fine-grained relative advantages \citep{feng2025gigpo}. Going beyond
trajectory-level rewards, GraphGPO aggregates rollout transitions into a
state-transition graph and assigns step-level credit according to graph
distances from intermediate states to the task goal
\citep{cheng2026graphgpo}. These advances substantially improve fine-grained
credit assignment without requiring an additional learned value model.

Despite this progress, existing graph-based credit assignment remains
\emph{batch-local}. At each policy update, the transition graph is constructed
only from trajectories sampled in the current rollout batch and discarded
afterward. Useful transitions discovered at different stages of training
therefore cannot be connected. For example, an early policy may discover a
promising prefix but fail before reaching the goal, whereas a later policy may
discover a successful continuation from an overlapping state. If the two
trajectory fragments occur in different policy updates, a batch-local graph
cannot recover their connectivity or propagate credit through the resulting
path. This limitation is particularly pronounced under small rollout groups,
where graph coverage is sparse and relative advantage estimates can be
sensitive to individual successful or failed trajectories.

A seemingly direct solution is to retain historical trajectories and replay
them during subsequent updates. However, trajectories generated by earlier
policies contain stale actions and log-probabilities. Directly including them
in the current policy loss can therefore introduce distribution mismatch.
Historical experience should instead improve the credit assigned to
current-policy rollouts without itself receiving gradients. This motivates the
central question of this work:

\begin{quote}
\emph{Can graph-based credit assignment accumulate experience across policy
updates while preserving optimization on current-policy trajectories?}
\end{quote}

To answer this question, we propose \emph{Temporal Instance-Graph Policy
Optimization} (TIGPO). TIGPO maintains a persistent transition graph indexed
by a stable task identity. During each policy update, valid historical
transitions associated with the same task are combined with transitions from
the current rollouts before distances to the goal and graph-based step
advantages are recomputed. This allows trajectory fragments discovered by
different policy versions to form complete paths to success and provide denser
credit signals for current behavior. Importantly, historical transitions are
used only as structural references: historical tokens, actions, and
log-probabilities are never included in the current policy loss.

Maintaining a persistent graph alone does not guarantee that the policy will
return to a previously explored task. TIGPO therefore introduces an
Exploration--Revisit rollout schedule that divides a fixed task-group budget
between ordinary exploration and scheduled revisitation. Exploration slots
follow the original task sampler, whereas Revisit slots reattempt previously
explored tasks after a controlled delay using the current policy. These
revisits actively reconnect current exploration with historical graph
structure, allowing the policy to extend or complete paths discovered during
earlier updates without increasing the per-update rollout budget.

A scheduled revisit also provides a task-matched cross-temporal comparison.
For each Revisit group, TIGPO combines its current episode-derived scores with
detached scores from the corresponding earlier Exploration group. The
resulting enlarged reference set is designed to stabilize relative advantage
estimation under small rollout groups. At the same time, comparing two policy
stages on the same task provides a direct signal of policy improvement. Only
the current Revisit trajectories receive gradients; the earlier scores serve
exclusively as detached reference statistics.

We evaluate TIGPO on two challenging long-horizon agent benchmarks, ALFWorld
\citep{shridhar2021alfworld} and WebShop \citep{yao2022webshop}. Across both
environments, TIGPO consistently outperforms prior group-based and graph-based
policy optimization methods. These results demonstrate that persistent
instance graphs, controlled historical revisitation, and cross-temporal group
comparison provide complementary benefits for long-horizon credit assignment.

Our main contributions are summarized as follows:

\begin{itemize}
    \item We introduce \textbf{cross-update persistent instance graphs} that
    connect valid transitions discovered by different policy versions and
    improve step-level credit assignment for current rollouts without
    replaying historical trajectories in the policy loss.

    \item We propose an \textbf{Exploration--Revisit rollout scheduler} that
    balances ordinary task exploration with delayed, task-specific
    revisitation under a fixed rollout budget, enabling the current policy to
    actively reuse and extend historical graph structure.

    \item We develop a \textbf{cross-temporal comparison mechanism} that pairs
    detached scores from an earlier Exploration group with the corresponding
    current Revisit group. It provides an enlarged reference set designed to
    stabilize relative advantage estimation while capturing policy
    improvement across training stages.
\end{itemize}

\section{Related Work}
\label{sec:related_work}

\paragraph{Reinforcement learning for LLM agents.}
LLMs are increasingly used as interactive agents that reason over observations
and execute actions in external environments. ReAct
\citep{yao2023react} establishes a general reasoning-and-acting paradigm,
while ALFWorld \citep{shridhar2021alfworld} and WebShop
\citep{yao2022webshop} provide representative benchmarks for embodied and web
agents. Their long interaction horizons and sparse outcome rewards make
fine-grained credit assignment particularly challenging. PPO
\citep{schulman2017ppo} estimates action advantages using a learned value
function, whereas GRPO \citep{shao2024deepseekmath} replaces the critic with
within-group reward normalization. GiGPO \citep{feng2025gigpo} introduces
episode-level and anchor-state-level groups for agent training, and GraphGPO
\citep{cheng2026graphgpo} derives step-level credit from a state-transition
graph constructed from the current rollout batch. In contrast, TIGPO jointly
introduces persistent task-conditioned graph memory, scheduled task
revisitation, and cross-temporal comparison, allowing experience collected at
different policy stages to improve credit assignment for current actions.

\paragraph{Historical experience reuse.}
Experience replay improves sample efficiency by retaining and resampling past
transitions \citep{schaul2016prioritized}, but directly optimizing on
trajectories generated by earlier policies introduces off-policy mismatch and
may require explicit correction \citep{espeholt2018impala}. TIGPO does not
replay historical trajectories in the policy objective. Historical
transitions are retained only as graph structure, while earlier rollout scores
serve as detached reference statistics. All trajectories receiving gradients
are newly generated by the current policy. TIGPO therefore reuses historical
structure and statistics without performing policy updates on stale actions
or log-probabilities.

\section{Preliminaries}
\label{sec:preliminaries}

\paragraph{Group-based policy optimization.}
Let $x\sim\mathcal{D}$ denote a task instance defining a finite-horizon
Markov decision process
$\mathcal{M}_x=(\mathcal{S}_x,\mathcal{A}_x,P_x,r_x)$.
At step $t$, an LLM policy $\pi_\theta$ samples an action
$a_{i,t}$ conditioned on the interaction history $h_{i,t}$ and observes
the next state $s_{i,t+1}$. A rollout
$\tau_i=(s_{i,0},a_{i,0},\ldots,s_{i,T_i})$ receives an episode outcome
$R_i$.

At each policy update, group-based policy optimization samples $B$ task
instances and generates $K$ rollouts for each task, giving a total rollout
budget of $N=BK$. The $K$ rollouts of the same task form a
\emph{task group}. For task $x$, let
$\mathcal{B}_x=\{\tau_i\}_{i=1}^{K}$ and
$\mathcal{Z}_x=\{R_i\}_{i=1}^{K}$ denote its rollout group and outcome
scores. GRPO \citep{shao2024deepseekmath} assigns each rollout a
within-group episode advantage
\begin{equation}
    \widehat{A}^{\mathrm{ep}}_i
    =
    \operatorname{Norm}(R_i;\mathcal{Z}_x),
    \qquad
    \operatorname{Norm}(z;\mathcal{Z})
    =
    \frac{z-\mu(\mathcal{Z})}
         {\sigma(\mathcal{Z})+\epsilon},
    \label{eq:episode_advantage}
\end{equation}
where $\epsilon>0$ ensures numerical stability. All actions in a rollout
inherit the same episode-level advantage.

\paragraph{Graph-based step-level credit.}
GraphGPO \citep{cheng2026graphgpo} further aggregates the rollout
transitions of task $x$ into a directed state-transition graph
$G_x=(V_x,E_x)$, where identical states are merged and valid transitions
form directed edges. Let $g_x$ be a successful terminal node and
$d_{G_x}(s,g_x)$ the shortest-path distance from state $s$ to success.
For a transition
$s_{i,t}\xrightarrow{a_{i,t}}s_{i,t+1}$, its graph return is
\begin{equation}
    Q^{G}_{i,t}
    =
    C\gamma^{d_{G_x}(s_{i,t+1},g_x)},
    \qquad 0<\gamma<1.
    \label{eq:graph_return}
\end{equation}
The successor state is used because it represents the direct consequence
of the current action; transitions leading closer to success therefore
receive larger returns. Returns of transitions sharing the same source
state are normalized to obtain the step-level graph advantage
$\widehat{A}^{G}_{i,t}$. GraphGPO then combines the episode- and
step-level signals as
\begin{equation}
    \widehat{A}_{i,t}
    =
    \lambda_{\mathrm{step}}\widehat{A}^{G}_{i,t}
    +
    \lambda_{\mathrm{ep}}\widehat{A}^{\mathrm{ep}}_i.
    \label{eq:graphgpo_advantage}
\end{equation}

Both the graph and the comparison statistics above are constructed from
the current rollout group. Consequently, transition structure discovered
in one policy update cannot inform later attempts at the same task. TIGPO
addresses this limitation by carrying task-level graph structure and
comparison information across policy updates.

\section{Method}

\begin{figure*}[t]
    \centering
    \includegraphics[
        width=\textwidth,
        trim={6pt 5pt 6pt 5pt},
        clip
    ]{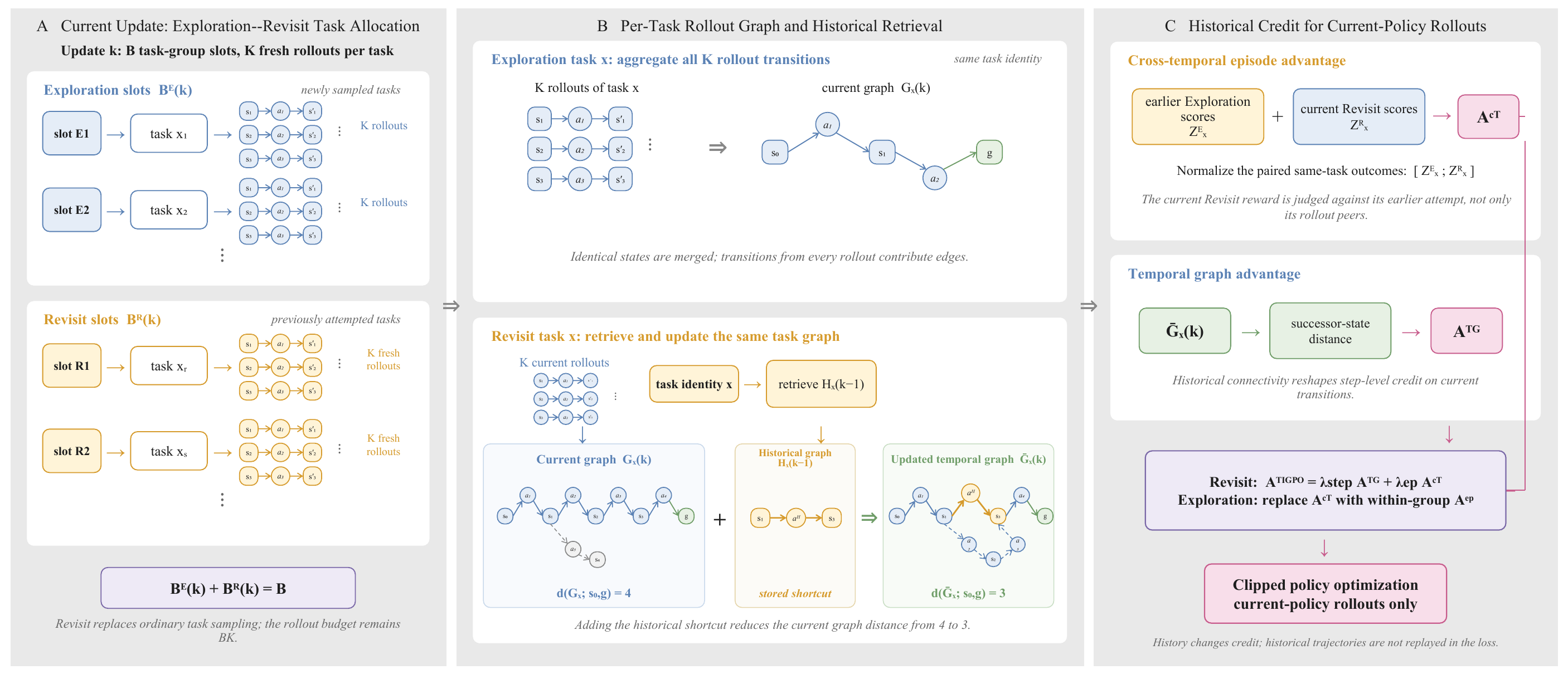}
    \caption{
        Overview of \textbf{Temporal Instance-Graph Policy Optimization
        (TIGPO)}.
        \textbf{(A)} At each policy update, a fixed budget of $B$ task
        groups is divided into Exploration and Revisit slots, with $K$
        fresh rollouts generated for each task.
        \textbf{(B)} Rollout transitions are aggregated into a per-task
        graph. For a revisited task, its historical graph is retrieved
        and combined with the current rollout graph, allowing previously
        discovered transitions to provide shorter routes to the goal.
        \textbf{(C)} The updated temporal graph provides step-level graph
        credit, while the paired Exploration and Revisit outcomes provide
        a cross-temporal episode advantage. Historical experience affects
        credit assignment, whereas the policy objective is optimized only
        on current-policy rollouts.
    }
    \label{fig:tigpo_framework}
\end{figure*}

Standard group-based policy optimization constructs supervision only
from the current rollout batch. As a result, useful transition structures
are discarded after each update, and repeated attempts at the same task
are evaluated independently. We propose \textbf{Temporal Instance-Graph
Policy Optimization (TIGPO)}, which uses the accumulated history of each
task to improve credit assignment for current-policy rollouts while
preserving on-policy optimization.

Figure~\ref{fig:tigpo_framework} summarizes the method.
As shown in Figure~\ref{fig:tigpo_framework}(A), TIGPO divides a fixed
budget of $B$ task groups into Exploration and Revisit slots, with $K$
fresh rollouts generated for each task. Exploration samples new task
attempts, whereas Revisit returns the current policy to previously
attempted tasks without increasing the total rollout budget $BK$.
In Figure~\ref{fig:tigpo_framework}(B), the rollouts of each task are
aggregated into a state-transition graph. For a revisited task, its
historical graph is retrieved and combined with the current graph,
allowing transition fragments discovered at different updates to form
shorter or previously unavailable routes to success.
Finally, Figure~\ref{fig:tigpo_framework}(C) converts this history into
two learning signals: a \emph{temporal graph advantage} derived from
successor-state distances, and a \emph{cross-temporal episode advantage}
obtained by comparing the current Revisit outcomes with their earlier
Exploration counterparts. TIGPO combines these signals to optimize only
fresh current-policy trajectories; historical experience affects credit
assignment but is never replayed in the policy loss.

\subsection{Temporal Instance Graph}
\label{sec:temporal_graph}

Long-horizon rollouts often expose only fragments of a useful solution. Under
a per-update graph, a prefix found in one update and a successful suffix found
later cannot support each other, even when they belong to the same task. This
motivates retaining task-conditioned transition structure across updates: the
graph should accumulate evidence about \emph{how states are connected}, while
policy optimization should remain on-policy.

\paragraph{Cross-update graph construction.}
TIGPO associates every task instance $x$ with a stable identity $\kappa(x)$.
Two attempts share experience only when they correspond to the same task
objective and initial environment configuration. This establishes a separate
learning history for each task while remaining independent of the particular
representation used by an environment.

Let $\mathcal{H}^{(k-1)}_x$ denote the transition graph accumulated for task
$x$ before policy update $k$, and let $G^{(k)}_{x}$ denote the graph formed by
its current rollouts. TIGPO constructs the temporal instance graph
\begin{equation}
    \overline G_x^{(k)}
    = \mathcal{H}^{(k-1)}_x \cup G^{(k)}_x.
    \label{eq:temporal_graph}
\end{equation}
The union connects compatible trajectory fragments discovered at different
stages of learning. For example, an earlier attempt may identify a useful
prefix, whereas a later attempt may connect a state on that prefix to the task
goal. Neither rollout batch needs to contain the complete path by itself.

\paragraph{Temporal graph credit.}
Shortest-path distances are recomputed on the temporal graph, and every
current transition receives
\begin{equation}
    q^{\mathrm{TG}}_{i,t}
    = C\gamma^{d_{\overline G_x^{(k)}}(s_{i,t+1},g_x)}.
    \label{eq:temporal_graph_return}
\end{equation}
Normalizing these returns among current transitions with the same source state
gives the temporal graph advantage $\widehat A^{\mathrm{TG}}_{i,t}$. After the
update, newly observed valid transitions are incorporated into
$\mathcal{H}^{(k)}_x$ for subsequent attempts. Historical transitions therefore
affect current credit through graph connectivity, but only current transitions
receive advantages and participate in policy optimization.

\subsection{Exploration--Revisit Sampling}
\label{sec:revisit_scheduling}

Accumulating a temporal graph alone does not ensure that the evolving policy
will return to the same task. When tasks are drawn independently from a broad
training distribution, such returns may be rare or occur only after an
uncontrolled delay. The graph would then store useful structure without
creating a timely opportunity to exploit it. This motivates coupling persistent
memory with an explicit mechanism for collecting a new on-policy attempt of a
previously explored task.

\paragraph{Fixed-budget scheduling.}
TIGPO partitions the $B$ task groups at update $k$ into Exploration and Revisit
groups:
\begin{equation}
    B_{\mathrm{E}}^{(k)}+B_{\mathrm{R}}^{(k)}=B.
    \label{eq:fixed_budget}
\end{equation}
Each group contains $K$ current-policy rollouts. An Exploration group draws a
task from the ordinary training distribution. The sampled task becomes
eligible for revisitation after a delay of $\Delta$ policy updates. A Revisit
group selects an eligible task and generates a new group of rollouts from the
current policy under the same task condition. If insufficient tasks are
eligible, the unused Revisit groups are reassigned to Exploration.

This schedule separates \emph{where to collect experience} from \emph{how much
experience to collect}. Revisit groups replace Exploration groups rather than
being appended to the batch, so every update still contains exactly $B$ groups
and $N=BK$ rollouts. Early updates emphasize discovering transition structure,
whereas later updates jointly expand that structure and reassess earlier tasks
with an improved policy.

\subsection{Cross-Temporal Policy Optimization}
\label{sec:cross_temporal}

Outcome normalization based only on a small current group can be unstable in
sparse-reward tasks: a single successful rollout may dominate the reference,
whereas uniformly failed rollouts provide little relative signal. A previous
attempt of the \emph{same} task offers a relevant additional baseline without
mixing outcomes from different task conditions. TIGPO therefore uses the
Exploration--Revisit pair not only to expand the graph but also to form a
temporally aligned comparison set.

\paragraph{Cross-temporal episode advantage.}
An Exploration--Revisit pair provides two outcome sets for the same task at
different stages of policy learning. Let $\mathcal{Z}^{\mathrm{E}}_x$ denote
the episode-derived scores from an Exploration group and
$\mathcal{Z}^{\mathrm{R}}_x$ those from its current Revisit group. The earlier
scores are treated as fixed reference values. For a current Revisit decision
with score $z^{\mathrm{R}}_{i,t}$, TIGPO defines the cross-temporal episode
advantage as
\begin{equation}
    \widehat A^{\mathrm{CT}}_{i,t}
    = \operatorname{Norm}\!\left(
        z^{\mathrm{R}}_{i,t};
        [\mathcal{Z}^{\mathrm{E}}_x;\mathcal{Z}^{\mathrm{R}}_x]
      \right),
    \label{eq:cross_temporal_advantage}
\end{equation}
where $[\cdot;\cdot]$ denotes concatenation. This comparison enlarges the
reference set available to a Revisit group and directly measures the relative
quality of the current attempt against an earlier policy on the same task.
Exploration groups retain the standard within-group episode advantage
$\widehat A^{\mathrm{ep}}_{i,t}$.

\paragraph{Joint policy objective.}
TIGPO combines episode-level comparison with temporal graph credit:
\begin{equation}
    \widehat A^{\mathrm{TIGPO}}_{i,t}
    = \lambda_{\mathrm{step}}\widehat A^{\mathrm{TG}}_{i,t}
    + \lambda_{\mathrm{ep}}
    \begin{cases}
        \widehat A^{\mathrm{ep}}_{i,t}, & \text{Exploration},\\
        \widehat A^{\mathrm{CT}}_{i,t}, & \text{Revisit}.
    \end{cases}
    \label{eq:tigpo_advantage}
\end{equation}
The resulting advantage is used in the clipped policy objective
\begin{equation}
    \mathcal{L}_{\mathrm{TIGPO}}(\theta)
    =-\mathbb{E}_{(i,t)\sim\mathcal{B}^{(k)}}
    \!\left[
      \min\!\left(
        \rho_{i,t}(\theta)\widehat A^{\mathrm{TIGPO}}_{i,t},
        \operatorname{clip}(\rho_{i,t}(\theta),1-\epsilon,1+\epsilon)
        \widehat A^{\mathrm{TIGPO}}_{i,t}
      \right)
    \right],
    \label{eq:tigpo_objective}
\end{equation}
where
$\rho_{i,t}(\theta)=
\pi_\theta(a_{i,t}\mid h_{i,t})/
\pi_{\theta_{\mathrm{old}}}(a_{i,t}\mid h_{i,t})$
and $\mathcal{B}^{(k)}$ contains only rollouts generated at update $k$.
Accordingly, past experience influences the current update through the
temporal graph and the cross-temporal reference distribution, without replaying
historical trajectories in the policy loss.

\section{Experiments}
\label{sec:experiments}

\subsection{Experimental Setup}
\label{sec:experimental_setup}

\paragraph{Benchmarks.}
We evaluate TIGPO on two challenging long-horizon interactive benchmarks:
ALFWorld \citep{shridhar2021alfworld} and WebShop
\citep{yao2022webshop}. ALFWorld is a text-based embodied environment in which
an agent completes household tasks through multi-step interactions. It
contains 3,827 task instances spanning six task categories: Pick, Clean, Cool,
Look, Heat, and Pick Two. We report the success rate for
each task category and the overall success rate.

WebShop is a web-based interactive environment that requires an agent to
search for and purchase products satisfying a natural-language instruction.
The agent must navigate product pages, select appropriate attributes, and make
a final purchase through a sequence of textual actions. We use the official
text-rich environment, which exposes information about interactive elements
such as search boxes, product options, and buttons. Following prior work, we
report both the average task score, which measures the degree to which the
selected product satisfies the instruction, and the success rate, which
requires complete satisfaction of the shopping goal.

\paragraph{Compared Methods.}
Following the comparison suite of \citet{cheng2026graphgpo}, we consider three
categories of baselines. First, we include closed-source general-purpose LLMs,
namely GPT-4o \citep{openai2024gpt4o} and Gemini-2.5-Pro
\citep{geminiteam2023gemini}. Second, we evaluate prompting-based agents,
including direct prompting with Qwen2.5 \citep{qwen2025qwen25}, ReAct
\citep{yao2023react}, which interleaves reasoning and acting, and Reflexion
\citep{shinn2023reflexion}, which uses verbal feedback from previous attempts
to refine subsequent decisions. These methods do not update model parameters
through reinforcement learning.

Third, we compare against representative RL-based methods: PPO
\citep{schulman2017ppo}, a critic-based policy optimization algorithm; RLOO
\citep{ahmadian2024backtobasics}, a REINFORCE-style method using a
leave-one-out baseline; GRPO \citep{shao2024deepseekmath}, which estimates
trajectory-level advantages from within-group reward statistics; GiGPO
\citep{feng2025gigpo}, which combines episode-level and anchor-state-level
relative advantages; and GraphGPO \citep{cheng2026graphgpo}, which derives
step-level credit from the state-transition graph constructed within the
current rollout batch. We locally reproduce GraphGPO using the same base
model, total rollout budget, training horizon, prompts, reward functions, and
evaluation protocol as TIGPO. Results adopted from prior work are marked with
\(\dagger\) in the result tables, whereas unmarked results are obtained using
our local evaluation pipeline.

\paragraph{Implementation Details.}
Following prior work, we use Qwen2.5-1.5B-Instruct
\citep{qwen2025qwen25} as the base model and retain the two most recent
interaction steps in the agent context. The model generates reasoning within
\texttt{<think>} tags followed by an action within \texttt{<action>} tags.
TIGPO and our reproduced GraphGPO baseline are trained for 250 policy updates
using the same prompts, reward functions, optimization settings, and total
rollout budget. The actor learning rate is \(1\times10^{-6}\), the KL-loss
coefficient is \(0.01\), and the rollout and evaluation temperatures are
\(1.0\) and \(0.4\), respectively. Successful episodes receive a reward of
\(10\), and invalid actions incur a penalty of \(-0.1\).

Both methods sample 64 current-policy trajectories per update. GraphGPO uses
eight task groups with eight trajectories per group, whereas TIGPO uses 16
groups with four trajectories per group. During temporally interleaved
training, TIGPO allocates eight groups to Exploration and eight to Revisit,
preserving eight newly sampled task groups per update without increasing the
rollout cost. The revisit delay is set to 10 updates, and cross-temporal
pairing is enabled for all eligible Revisit groups. The step-level and
episode-level advantage weights are both set to \(1\), with distance-discount
factors of \(0.10\) for ALFWorld and \(0.20\) for WebShop. We evaluate the
checkpoint at update 250 over 512 episodes for each of three seeds
(\(123,456,789\)) and report the mean and standard deviation. 

\begingroup
\newcommand{\res}[2]{%
  \ensuremath{#1_{\scriptscriptstyle(#2)}}%
}
\newcommand{\bres}[2]{%
\ensuremath{\mathbf{#1}_{\scriptscriptstyle(\mathbf{#2})}}%
}

\begin{table}[t]
\centering
\caption{
The test performance on ALFWorld and WebShop.
For ALFWorld, we report the average success rate (\%) for each subtask
as well as the overall result.
For WebShop, we report the average task score and the average success
rate (\%).
Most results are averaged over 3 random seeds during testing.
The best performances are highlighted in bold.
}
\label{tab:main_results}

\setlength{\tabcolsep}{3.4pt}
\renewcommand{\arraystretch}{1.02}

\resizebox{\textwidth}{!}{%
\begin{tabular}{ll|ccccccc|cc}
\toprule
\textbf{Type}
& \textbf{Method}
& \multicolumn{7}{c|}{\textbf{ALFWorld}}
& \multicolumn{2}{c}{\textbf{WebShop}}                       \\
\cmidrule(lr){3-9}
\cmidrule(lr){10-11}
&
& Pick
& Clean
& Cool
& Look
& Heat
& Pick2
& All
& Score
& Succ.                                                       \\
\midrule

\multicolumn{11}{l}{\textbf{\emph{Closed-Source Models}}}    \\

Prompting
& GPT-4o
& 75.3 & 60.8 & 31.2 & 56.7 & 21.6 & 49.8 & 48.0
& 31.8 & 23.7                                                \\

Prompting
& Gemini-2.5-Pro
& 92.8 & 63.3 & 62.1 & 69.0 & 26.6 & 58.7 & 60.3
& 42.5 & 35.9                                                \\

\midrule
\multicolumn{11}{l}{\textbf{\emph{Qwen2.5-1.5B-Instruct}}}   \\

Prompting
& Qwen2.5
& 5.9 & 5.5 & 3.3 & 9.7 & 4.2 & 0.0 & 4.1
& 23.1 & 5.2                                                 \\

Prompting
& ReAct
& 17.4 & 20.5 & 15.7 & 6.2 & 7.7 & 2.0 & 12.8
& 40.1 & 11.3                                                \\

Prompting
& Reflexion
& 35.3 & 22.2 & 21.7 & 13.6 & 19.4 & 3.7 & 21.8
& 55.8 & 21.9                                                \\

RL Training
& PPO
& \res{64.8}{3.5}
& \res{40.5}{6.9}
& \res{57.1}{4.9}
& \res{60.6}{6.6}
& \res{46.4}{4.0}
& \res{47.4}{1.9}
& \res{54.4}{3.1}
& \res{73.8}{3.0}
& \res{51.5}{2.9}                                            \\

RL Training
& RLOO
& \res{88.3}{3.0}
& \res{52.8}{8.6}
& \res{71.0}{5.9}
& \res{62.8}{8.7}
& \res{66.4}{5.5}
& \res{56.9}{4.7}
& \res{69.7}{2.5}
& \res{73.9}{5.6}
& \res{52.1}{6.7}                                            \\

RL Training
& GRPO
& \res{82.89}{3.62}
& \res{82.14}{6.37}
& \res{73.86}{6.84}
& \res{78.57}{0.00}
& \res{77.78}{4.54}
& \res{71.43}{3.89}
& \res{77.86}{1.33}
& \res{84.73}{0.49}
& \res{71.35}{2.05}                                          \\

RL Training
& GiGPO
& \res{97.38}{1.85}
& \res{90.48}{1.89}
& \bres{92.94}{2.66}
& \res{80.22}{1.10}
& \res{98.41}{0.40}
& \res{83.03}{1.94}
& \res{91.02}{0.32}
& \res{87.94}{0.43}
& \res{73.83}{2.30}                                          \\

RL Training
& GraphGPO
& \res{92.56}{3.07}
& \bres{92.18}{2.65}
& \res{88.97}{2.90}
& \res{86.29}{0.76}
& \bres{100.0}{0.00}
& \res{74.13}{7.13}
& \res{89.32}{1.83}
& \res{87.34}{1.09}
& \res{76.37}{0.32}                                         \\

\textbf{RL Training}
& \textbf{TIGPO}
& \bres{98.01}{0.9}
& \res{90.05}{1.38}
& \res{92.39}{1.31}
& \bres{89.63}{3.15}
& \res{90.50}{1.06}
& \bres{83.43}{3.90}
& \bres{91.28}{1.04}
& \bres{88.65}{2.30}
& \bres{77.54}{2.41}                                         \\

\bottomrule
\end{tabular}%
}
\end{table}
\endgroup

\subsection{Experimental Results}
\paragraph{Performance on Agentic Benchmarks.}

Table~\ref{tab:main_results} summarizes the results on ALFWorld and
WebShop. Under the same rollout budget, TIGPO achieves the best aggregate
performance on both benchmarks. On ALFWorld, TIGPO obtains an overall
success rate of \(91.28\%\), outperforming GiGPO and GraphGPO by \(0.26\)
and \(1.96\) percentage points, respectively. TIGPO achieves the best
performance on three of the six task categories, including Pick, Look,
and Pick2. The improvements are particularly pronounced on Look and
Pick2, where TIGPO surpasses GraphGPO by \(3.34\) and \(9.30\) percentage
points, respectively. Although GiGPO and GraphGPO remain stronger on
several individual task categories, TIGPO provides the best overall
balance across the diverse ALFWorld tasks.

TIGPO also consistently improves the aggregate WebShop metrics. It
achieves an average task score of \(88.65\) and a success rate of
\(77.54\%\), outperforming GiGPO by \(0.71\) and \(3.71\) percentage
points and GraphGPO by \(1.31\) and \(1.17\) percentage points,
respectively. These improvements are obtained without increasing the
per-update rollout budget. The results suggest that retaining
task-specific transition structure across policy updates, actively
revisiting previously explored tasks, and introducing cross-temporal
references allow TIGPO to exploit experience beyond the current rollout
batch, leading to stronger aggregate performance across both embodied
and web-based agent environments.


\begingroup

\newcommand{\res}[2]{%
  \ensuremath{#1_{\scriptscriptstyle(#2)}}%
}

\newcommand{\bres}[2]{%
  \ensuremath{\mathbf{#1}_{\scriptscriptstyle(\mathbf{#2})}}%
}

\begin{table}[t]
    \centering
    \caption{
        Ablation results on ALFWorld.
        \emph{Current-update graph} constructs graph structure
        independently within each policy update.
        \emph{Persistent Graph} retains task-specific graph memory
        across updates but disables Exploration--Revisit scheduling
        and cross-temporal comparison.
        \emph{Persistent Graph + Exploration--Revisit} introduces
        scheduled revisitation while retaining batch-local advantage
        normalization.
        Full TIGPO additionally introduces cross-temporal comparison.
        All variants use 64 trajectories per policy update.
        We report mean success rates (\%) over three evaluation seeds,
        with standard deviations shown in parentheses.
        The best result in each column is highlighted in bold.
    }
    \label{tab:ablation_alfworld}

    \setlength{\tabcolsep}{4.2pt}
    \renewcommand{\arraystretch}{1.08}

    \resizebox{\textwidth}{!}{%
    \begin{tabular}{l|ccccccc}
        \toprule
        \textbf{Method}
        & \multicolumn{7}{c}{\textbf{ALFWorld}} \\
        \cmidrule(lr){2-8}
        & Pick
        & Clean
        & Cool
        & Look
        & Heat
        & Pick2
        & All \\
        \midrule

        Current-update graph
        & \res{92.56}{3.07}
        & \res{92.18}{2.65}
        & \res{88.97}{2.90}
        & \res{86.29}{0.76}
        & \bres{100.00}{0.00}
        & \res{74.13}{7.13}
        & \res{89.32}{1.83} \\

        Persistent Graph
        & \res{94.39}{1.66}
        & \bres{95.17}{1.49}
        & \res{89.73}{2.13}
        & \res{89.63}{3.15}
        & \res{92.61}{3.06}
        & \res{80.69}{3.05}
        & \res{90.69}{0.49} \\

        Persistent Graph + Exploration--Revisit
        & \res{94.98}{0.74}
        & \res{88.52}{2.73}
        & \res{87.13}{3.67}
        & \bres{95.46}{1.89}
        & \res{97.82}{1.56}
        & \res{78.72}{1.08}
        & \res{89.58}{0.45} \\

        \textbf{TIGPO}
        & \bres{98.01}{0.90}
        & \res{90.05}{1.38}
        & \bres{92.39}{1.31}
        & \res{89.63}{3.15}
        & \res{90.50}{1.06}
        & \bres{83.43}{3.90}
        & \bres{91.28}{1.04} \\

        \bottomrule
    \end{tabular}%
    }
\end{table}

\endgroup

\paragraph{Ablation Study.}
Table~\ref{tab:ablation_alfworld} examines the contribution of
persistent task-specific graph memory, Exploration--Revisit
scheduling, and cross-temporal comparison. Starting from a graph
constructed independently within each policy update, carrying graph
structure across updates improves the overall success rate from
\(89.32\%\) to \(90.69\%\). The Persistent Graph variant improves
five of the six task categories, with particularly clear gains of
\(3.34\) percentage points on Look and \(6.56\) points on Pick2.
These results demonstrate that valid transitions discovered during
earlier policy updates provide useful structural information for
credit assignment in current rollouts.

Adding Exploration--Revisit scheduling without cross-temporal
comparison results in an overall success rate of \(89.58\%\).
Although scheduled revisitation substantially improves performance
on Look and maintains strong performance on Heat, it reduces
performance on several other task categories. This result indicates
that revisiting previously explored tasks alone is insufficient.
Because the fixed rollout budget is distributed across more task
groups, each group contains fewer trajectories, making batch-local
relative advantage estimation less informative.

Full TIGPO addresses this limitation by pairing each Revisit group
with detached scores from its corresponding earlier Exploration
group. Introducing cross-temporal comparison improves the overall
success rate by \(1.70\) percentage points over the variant without
cross-temporal comparison, reaching the best overall performance of
\(91.28\%\). TIGPO also achieves the strongest results on Pick, Cool,
and Pick2. While no single variant dominates every task category,
full TIGPO provides the best aggregate performance, improving the
current-update graph baseline by \(1.96\) percentage points. These
results show that persistent graph memory and cross-temporal
comparison play complementary roles: the former preserves structural
experience across updates, while the latter provides a more
informative reference for optimizing newly generated trajectories.


\begin{figure}[t]
    \centering
    \includegraphics[
        width=\textwidth
    ]{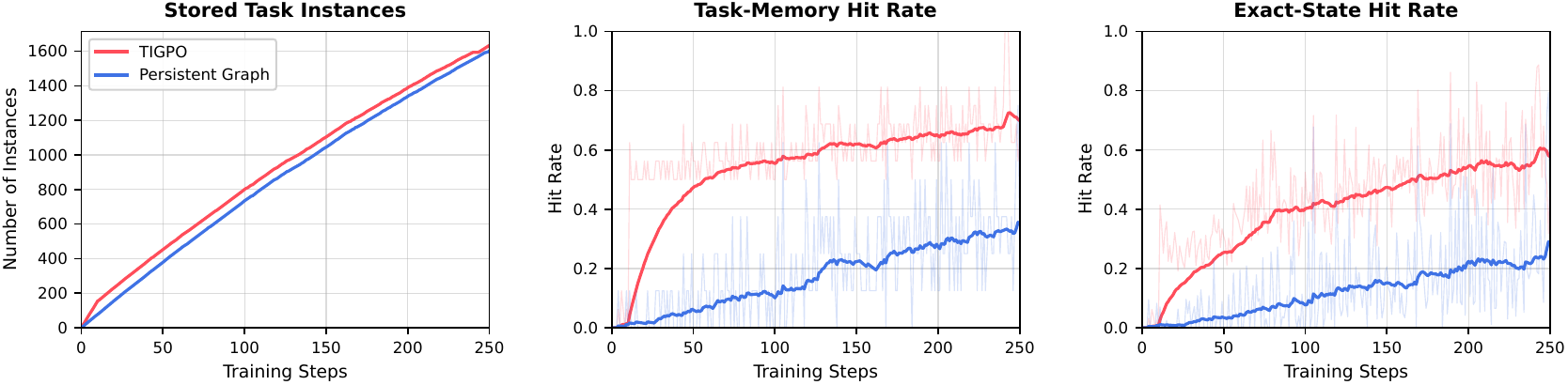}

    \caption{
        Historical experience utilization on ALFWorld.
        From left to right, the panels show the number of stored task
        instances, the task-memory hit rate, and the exact-state hit
        rate. Lighter curves show the original observations, while
        darker curves show exponential moving averages with a decay
        of \(\alpha=0.95\).
    }
    \label{fig:history_utilization}
\end{figure}

\paragraph{Historical Experience Utilization.}
To understand how the temporal design affects the use of persistent
graph memory, Figure~\ref{fig:history_utilization} compares full
TIGPO with the Persistent Graph ablation from
Table~\ref{tab:ablation_alfworld}. Both variants retain task-specific
transition graphs across policy updates, but the Persistent Graph
ablation disables Exploration--Revisit scheduling and cross-temporal
comparison.

The two variants accumulate a comparable number of task-specific
graph records throughout training. However, TIGPO achieves
substantially higher task-memory and exact-state hit rates. The
higher task-memory hit rate indicates that current rollout groups
more frequently correspond to tasks for which useful historical
transitions are available. The higher exact-state hit rate further
shows that the retrieved graph structure overlaps with the states
actually encountered by the current policy, rather than merely
belonging to the same task instance.

These observations indicate that accumulating persistent graph
structure alone does not ensure that the stored experience will be
encountered again during subsequent policy updates.
Exploration--Revisit scheduling actively reconnects current rollouts
with previously explored tasks, increasing the opportunity to reuse
relevant historical structure. Together with the ablation results,
these diagnostics support the role of TIGPO's temporal design as an
active mechanism for making persistent graph memory accessible to
current-policy credit assignment.

\paragraph{Computational Overhead.}
Table~\ref{tab:computational_efficiency} compares TIGPO with GraphGPO on
ALFWorld using the same model, hardware, and per-update rollout budget of 64
trajectories. We discard the first 10 updates as warm-up and subtract the
separately logged evaluation time from the end-to-end step time. We report the
median and interquartile range (IQR) over the remaining updates to reduce the
influence of periodic runtime spikes. The end-to-end timer encloses the entire
training iteration and therefore also captures the cost of TIGPO's history
maintenance, Exploration--Revisit scheduling, and cross-temporal processing,
even though these operations are not timed separately.

TIGPO requires 247.08 seconds per training update, compared with 258.32 seconds
for GraphGPO, and its core LLM computation time remains comparable
(59.56 vs.\ 60.42 seconds). Its peak allocated GPU memory is also effectively
unchanged (84.31 vs.\ 84.36 GB). The corresponding component-wise median times
for TIGPO and GraphGPO are 38.84 vs.\ 39.90 seconds for actor update, 10.56
vs.\ 10.65 seconds for old log-probability computation, 10.05 vs.\ 10.32 seconds
for the reference-model forward pass, and 0.339 vs.\ 0.342 seconds for reward
computation. Since the IQRs largely overlap, we interpret the small timing
differences as comparable runtime rather than as evidence of speedup. Together
with the performance gains in Table~\ref{tab:main_results}, these results show
that TIGPO improves agentic performance without introducing observable
end-to-end training-time or GPU-memory overhead relative to GraphGPO.
\begin{table*}[t]
    \centering
    \caption{Computational efficiency on ALFWorld. Values are median (IQR)
    over training updates 11--250. Evaluation time is excluded from training
    step time. Core LLM compute is the per-step sum of actor update, old
    log-probability computation, and reference-model computation. Both methods
    use 64 trajectories per update.}
    \label{tab:computational_efficiency}
    \small
    \setlength{\tabcolsep}{4pt}
    \begin{tabular}{lccc}
        \toprule
        Method
        & Step time (s) $\downarrow$
        & Core LLM (s) $\downarrow$
        & GPU memory (GB) $\downarrow$ \\
        \midrule
        GraphGPO
        & 258.32 (225.71--316.23)
        & 60.42 (45.07--82.03)
        & 84.36 (83.44--84.43) \\
        TIGPO
        & \textbf{247.08} (226.00--313.07)
        & \textbf{59.56} (46.85--85.47)
        & \textbf{84.31} (83.51--84.46) \\
        \midrule
        Relative change
        & $-4.35\%$
        & $-1.43\%$
        & $-0.06\%$ \\
        \bottomrule
    \end{tabular}
\end{table*}

\section{Conclusion}

We introduced TIGPO, a graph-based policy optimization framework that extends agent learning beyond trajectory-local and single-update credit assignment. TIGPO maintains a persistent task-specific transition graph across policy updates, schedules Exploration--Revisit rollouts to reuse previously discovered experience, and constructs cross-temporal reference groups for more informative advantage estimation. Experiments on ALFWorld and WebShop demonstrate that TIGPO consistently improves over GRPO, GiGPO, and GraphGPO, achieving strong performance across both embodied decision-making and web-based interaction tasks. The ablation results further confirm that persistent graph memory, scheduled revisitation, and cross-temporal comparison provide complementary benefits. Moreover, TIGPO introduces no observable end-to-end training-time or GPU-memory overhead relative to GraphGPO under the same rollout budget. These results suggest that preserving and reusing structured experience across policy updates is an effective and computationally practical direction for training capable LLM agents. Future work may explore scaling TIGPO to larger models and more diverse environments, as well as developing adaptive graph compression and retrieval strategies for longer training horizons.

\clearpage
\bibliography{iclr2027_conference}
\bibliographystyle{iclr2027_conference}


\end{document}